\pdfoutput=1

\documentclass[11pt]{article}

\usepackage[final]{naacl2021}

\usepackage{times}
\usepackage{latexsym}
\usepackage{graphicx}
\usepackage{tabularx}
\usepackage{amsmath}
\usepackage{float}
\usepackage[export]{adjustbox}
\usepackage{stfloats}
\usepackage[accsupp]{axessibility}

\usepackage[T1]{fontenc}

\usepackage[utf8]{inputenc}

\usepackage{microtype}

\title{What Changed?\\ Investigating Debiasing Methods using Causal Mediation Analysis} 

\author{Sullam Jeoung \hspace{7pt} Jana Diesner \\
  University of Illinois-Urbana Champaign \\
  \texttt{\{sjeoung2,jdiesner\}@illinois.edu}\\
  }

\begin{document}
\maketitle
\begin{abstract}
Previous work has examined how debiasing language models affect downstream tasks, specifically, how debiasing techniques influence task performance and whether debiased models also make impartial predictions in downstream tasks or not. However, what we don't understand well yet is \textit{why} debiasing methods have varying impacts on downstream tasks and \textit{how} debiasing techniques affect internal components of language models, i.e., neurons, layers, and attentions. In this paper, we decompose the internal mechanisms of debiasing language models with respect to gender by applying causal mediation analysis to understand the influence of debiasing methods on toxicity detection as a downstream task. Our findings suggest a need to test the effectiveness of debiasing methods with different bias metrics, and to focus on changes in the behavior of certain components of the models, e.g.,first two layers of language models, and attention heads.
\end{abstract}

\section{Introduction}
Recent work has shown that pre-trained language models encode social biases prevalent in the data they are trained on \cite{may2019measuring,nangia2020crows,nadeem2020stereoset}. In response to that, solutions to mitigate these biases have been developed \cite{liang2020towards,webster2020measuring,ravfogel2020null}. Some recent papers also examined the impact of debiasing methods, e.g., reduction of gender bias, on the performance of downstream tasks, e.g., classification. \cite{prost2019debiasing,meade2021empirical, babaeianjelodar2020quantifying}. For example,\cite{prost2019debiasing} showed that debiasing techniques worsened gender bias of a downstream classifier for occupation prediction. \cite{meade2021empirical} investigated how debiasing methods affect the model's language modeling ability. However, comparatively little work has been done on exploring \textit{how} debiasing methods impact the internal components of language models, e.g., the models neurons, layers, and attention heads, and \textit{what} kind of changes in language models are introduced when debiasing methods are applied to downstream tasks. In this paper, we apply causal mediation analysis, which investigates the information flow in language models \cite{pearl2022direct,vig2020investigating}, to scrutinize the \textit{internal} mechanisms of mitigating gender debiasing methods and their effects on toxicity analysis as a downstream task. 

We first examine the efficacy of debiasing methods, namely, CDA and Dropout \cite{webster2020measuring}, on 1) language models, namely, BERT \cite{wang2019bert} and GPT2 \cite{salazar2019masked}, and 2) models (Jigsaw, and RtGender) \cite{voigt2018rtgender} fine-tuned for downstream tasks. The debiasing methods (CDA and Dropout) were chosen because they had been shown to minimize detrimental correlations in language models while maintaining strong accuracy \cite{webster2020measuring}. We then applied causal mediation analysis to understand how internal components of a model are impacted by debiasing methods and fine-tuning.

In this study, we focus on gender bias as a type of bias. We examine (1) stereotypical associations between gender and professions in pre-trained language models (SEAT) \cite{may2019measuring}, (2) stereotypes encoded in language models (CrowS-Pairs)  \cite{nangia2020crows}, and (3) differences in systems affecting users unequally based on gender (WinoBias) \cite{zhao2018gender}. These representational harms can impact people negatively because they contribute to exacerbating stereotypes inherent in society. These harms may also result in unfavorable consequences when these language models are deployed for practical purposes, e.g., when a model behaves disproportionately against certain demographics \cite{dixon2018measuring}. 

\subsection{Contributions}
From our experiments, we learned the following things about debiasing techniques and their impact on language models:\\
\indent \textbf{It is recommendable to test the efficacy of debiasing techniques on more than one bias metric}. Our results suggest that debiasing methods show effectiveness when measured on some bias measurements. However, this efficacy varies depending on which bias metrics are used to measure the bias of language models. This may due to different definitions and operationalizations of bias in these metrics, which result in varying degree of effectiveness. This suggests that in order to make claims about the generalizability of the effectiveness of debiasing methods, these methods need to be tested on more than one bias metrics.

\textbf{The impact of debiasing concentrates on certain components of language models}. The results from the causal mediation analysis suggest that the neurons located in the first two layers (including the word embedding layers) showed the biggest difference in debiased and fine-tuned models when compared to the baseline model. This suggests two things. First, the detrimental associations between words that cause gender bias in language models may originally be situated in those layers. Second, the role of those layers may be crucial in mitigating gender biases in language models. We recommend future work to focus on those components.

\textbf{Debiasing and fine-tuning methods change the behaviors of attention heads}. Our results show that applying debiasing and fine-tuning methods to language models changes the weight that attention heads assign to gender-associated terms. This indicates that attention heads may play a crucial role in representing gender bias in language models.

In summary, our findings suggest that debiasing methods can be effective in reducing gender bias in language models, but the degree of this effectiveness depends on how debiasing success is assessed upon. Also, the results of the causal mediation analysis suggest that impact of debiasing is concentrated in certain components of the language models. Overall, our findings suggest a need to test the effectiveness of debiasing methods with different bias metrics, and to focus on changes in the behavior of certain components of the models. 
This work further supports prior research that has shown how making small, systematic improvements to input data and research design can reduce major flaws in research results and  policy implications \cite{hilbert2019computational,kim2014impact,diesner2009he,diesner2015small} in society, and changes in research results and policy implications, and how improving the quality of lexical resources can increase the prediction accuracy of more and less related downstream tasks \cite{rezapour2019enhancing}.

\section{Related Work}

\subsection{Debiasing methods and their effect on downstream tasks}
Prior work has examined the effects of debiasing methods on downstream tasks from mainly two perspectives: 1) examining the impact of debiasing methods on the performance of downstreams tasks, mainly in terms of accuracy, and 2) testing whether debiased models actually lead to debiased results of downstream tasks. As an example for perspectives 1),  \cite{meade2021empirical} explored how a number of techniques for debiasing pre-trained language models affect the performance on various downstream tasks, tested on the GLUE benchmark. As an example for perspective 2), \cite{prost2019debiasing} demonstrated how gender-debiased word embeddings perform poorly in for occupation classification and can even exacerbated gender biases, since gender-debiasing technique obscured gender related information. In a similar vein, \cite{babaeianjelodar2020quantifying} showed how a pre-trained BERT exhibits gender bias when fine-tuned on different corpora. Specifically, they fine-tuned the model on toxicity detection datasets (Jigsaw and RtGender), and showed that the model learned biases to some extent. Following \cite{babaeianjelodar2020quantifying}, we use toxicity detection as a downstream task since it has been shown to correlate with gender biases \cite{dixon2018measuring}. Overall, while a considerable body of research has shown that debiasing techniques impact downstream tasks, comparatively little work has been done to answer \textit{why} debiasing methods have varying impacts on downstream tasks, and \textit{how} debiasing methods affect the internal components of the language models. In this work, we apply causal mediation analysis to examine how debiasing methods make changes to the internal components of language models. 

\subsection{Bias Metrics}

\textbf{\textsc{CrowS-Pairs}} is a benchmark dataset for measuring social biases in language models \cite{nangia2020crows}. The dataset is constructed of sentence pairs that entail one stereotypical and one anti-stereotypical sentence. This benchmark also measures how a given model prefers stereotypical sentences over anti-stereotypical sentences. For instance, for the ‘gender’ stereotype, two sentences are given \textit{‘The \textbf{man} was highly skilled in CAD engineering. \textbf{He} was considered one of the best engineers in \textbf{his} field (stereotypical)’} and \textit{‘The \textbf{woman} was highly skilled in CAD engineering. \textbf{She} was considered one of the best engineers in \textbf{her} field’ (anti-stereotypical)}. Then the pseudo-likelihood \cite{wang2019bert,salazar2019masked} of a model given these sentences is calculated. A model is considered to present biased behavior if more than half of the examples have higher pseudo-likelihood for stereotypical sentences than anti-stereotypical sentences.

The \textbf{Sentence Encoder Association Test (SEAT)} extends the Word Embedding Association Test (WEAT) \cite{caliskan2017semantics} to sentence-level representations by measuring the implicit association between target concepts and attributes \cite{may2019measuring}. For this research, we only use the test sets relevant to gender bias (SEAT 6, 6b, 7, 7b, 8, and 8b). Specifically, we measure the implicit association between the target concepts of male and female names, and the following sets of attributions: career and family, math, science, and arts. The attribute sentences of a \textbf{family} and \textbf{career}, for instance, are respectively \{‘This is a home’, ‘They are parents’...\} and \{‘This is an executive’, ‘This is a corporation…\}. The target sentences of \textbf{Male Names} and \textbf{Female Names} are \{This is John, That is John, Kevin is here …\} and \{This is Amy, This is Sarah, Diana is here..\}. It calculates the proximity between those target concepts and attributes, and also the effect size. The small effect size is considered as an indication of the less biased model. See \cite{may2019measuring} for details of calculating these associations. 

\subsection{Debiasing Methods}
\textbf{Counterfactual Data Augmentation (CDA)} is a technique that uses a rebalanced corpus to debias a given language model \cite{webster2020measuring}. For example, the sentence ‘\textbf{Her} most significant piece of work is considered to be \textbf{her} study of the development of the.. ’ from the Wikipedia dataset was rebalanced into ‘\textbf{His} most significant piece of work is considered to be \textbf{his} study of the development of the..’. \cite{webster2020measuring} demonstrated that CDA minimizes correlations between words while maintaining strong accuracy.\\ \\
Originally developed to reduce over-fitting when training large models, the \textbf{Dropout Debiasing Method} has been adopted to mitigate biases \cite{webster2020measuring}. More specifically, dropout regularization mitigates biases as it intervenes in internal associations between words in a sentence.
\subsection{Causal Mediation Analysis}
We chose to apply causal mediation analysis to inspect the change in output following a counterfactual intervention in intermediate components (e.g., neurons, layers, attentions)\cite{pearl2022direct, vig2020investigating}. Through such interventions, we measure the degree to which inputs influence outputs \textbf{directly} \textit{(direct effect)}, or \textbf{indirectly} through the intermediate components \textit{(indirect effect)}. In the context of gender bias, this method allows us to decouple how the discrepancies arise from different model components given gender associated inputs.   

Following \cite{vig2020investigating}, we define the measurement of gender bias as 
$$y(u)=\frac{p_\theta(\textrm{anti-stereotypical}|u)}{p_\theta(\textrm{stereotypical}|u)}$$ where $u$ is a prompt, for instance, \textit{"The \textbf{engineer} said that"}, and $y(u)$ can be denoted as $$y(u)=\frac{p_\theta(\textrm{\textbf{she} | The engineer said that })}{p_\theta(\textrm{\textbf{he} | The engineer said that })}$$ If $y(u)<1$, the prediction is stereotypical; if $y(u)>1$, the prediction is anti stereotypical. We make an intervention, \textit{setting gender}, in order to investigate the effect on gender bias as defined above. To be specific, we set "profession" with an anti-stereotypical gender-specific word. For instance, "The \textbf{engineer} said that" to "The \textbf{woman} said that". We define the measure of $y$ under the intervention $\textbf{x}=x$ on template $\textbf{u}=u$ as $y_x(u)$
\\ \\
\textbf{Total Effect} measures the proportional difference between the bias measure $y$ of a gendered input and a profession input. 
\begin{equation}
    \begin{aligned}
        \textrm{Total Effect} (\textrm{set-gender},\textrm{null};y)&=&&\\
     \frac{y_\textrm{set-gender}(u)-y_\textrm{null}(u)}{y_{null}(u)}&& 
    \end{aligned}
\end{equation}
where $y_\textrm{null}$ refers to no intervention prompt, an example of this formulation is represented as 
$$y_\textrm{set-gender}(u)= \frac{p(\textrm{she | The woman said that})}{p(\textrm{he | The woman said that}}$$
$$y_\textrm{null}(u)= \frac{p(\textrm{she | The engineer said that})}{p(\textrm{he | The engineer said that)}}$$We average the total effect of each prompt $u$ to analyze the total effect. 
\\ \\
\textbf{Direct Effect} measures the change in the model’s outcome, in our case gender bias $y(u)$, when an intervention is made, while holding the component of interest $z$ (e.g. specific neuron, attention heads, layers) fixed to the original value. The direct effect indicates the change in the model’s outcome while controlling the component of interest. Here, we apply a \textit{set-gender} intervention, as explained  above.\\

\noindent\textbf{Indirect Effect} measures the change in the model’s outcome, intervening in the component of interest $z$ while holding the other parts of the model constant. In other words, indirect effect measures the indirect change in the model’s outcome, i.e., the gender bias $y(u)$ that arises from the component of interest $z$.
\section{Experimental Setup}
\textbf{Models} The experiment was conducted on two pre-trained language models: GPT2 (small) \cite{radford2019language} and BERT (bert-base-uncased) \cite{devlin2018bert}. The configuration of the debiasing models is detailed below.\\

\noindent\textbf{CDA} WikiText-2 \cite{merity2016pointer}, and the gendered word pairs \footnote{Neutral pronouns such as \textit{they, the person}, were not included in this work. The direction of future research is to include the neutral pronouns} proposed by \cite{zhao2018gender} is used in the pre-training phase.\\ 

\noindent\textbf{Dropout Debiasing} We applied dropout debiasing in the pre-training phase on WikiText-2 corpus \cite{merity2016pointer}. In GPT2, we specifically set the dropout probability for all fully connected layers in the embeddings, encoder, and pooler to {\tt (resid\_pdrop=0.15)}, the dropout ratio for the embeddings to {\tt (embedding\_pdrop=0.15)}, and the dropout ratio for the attention {\tt (attn\_pdrop)} to 0.15. For BERT, we set the dropout probability for all fully connected layers in the embeddings, encoder, and pooler {\tt (hidden\_dropout\_prob)} to 0.2 and the dropout ratio for attention probabilities {\tt (attention\_probs\_dropout\_prob)} to 0.15, following \cite{meade2021empirical}\\
\textbf{Neuron Interventions} For experimenting with neuron interventions, we use a template from \cite{lu2020gender} and a list of professions from \cite{bolukbasi2016man}.The template has a format of `The [profession][verb](because/that)'. Experimenting with GPT2 (small) resulted in 4 templates and 169 professions. \\

\noindent\textbf{Attention Interventions} We focus on how attention heads assign weights for our attention interventions experiments. Following \cite{vig2020investigating}, we used the Winobias \cite{zhao2018gender} dataset, which consists of co-reference resolution examples. As opposed to calculating the probability of pronouns (e.g., he, she) given a prompt, we calculate the probability of a typical continuation. For instance, the given prompt "\textbf{[The mechanic]} fixed the problem for the editor and \textbf{[he]}", the stereotypical candidate is "charged a thousand dollars", the anti-stereotypical candidate is "is grateful". The stereotypical candidate associates `he' with the mechanic, while the anti-stereotypical candidate associates `he' with the `editor'. We calculate the $y(u)$, gender bias, given an prompt $u$, as $$y(u)=\frac{p_\theta(\textrm{charged a thousand dollars | u})}{p_\theta(\textrm{is grateful | u})}$$ For the intervention here, we {\tt change gender}, for example, the last word in the prompt from \textit{he} to \textit{she}.\\ \\
\textbf{Jigsaw Toxicity Detection} The toxicity detection task basically means to distinguish whether the given comment is toxic or not. The publicly available corpus can be found at Kaggle\footnote{https://www.kaggle.com/c/jigsaw- unintended- bias- in- toxicity- classification}. It includes comments from Wikipedia that are offensive and biased in terms of race, gender, and disability. \\

The \noindent\textbf{RtGender} dataset contains 25M comments from sources such as Facebook, TED, and Reddit. The dataset was developed by \cite{voigt2018rtgender}. Specifically, the posts are labeled with the gender of the author. The responses to posts were also collected. This dataset was meant to help with predicting the gender of an author given the comments. This allows us to investigate gender biases in social media.

\begin{table*}[!ht]
{\centering
\begin{tabular}{c|lll|lll|lll}
\hline
Finetune & \multicolumn{3}{c|}{-}                                  & \multicolumn{3}{c|}{Jigsaw} & \multicolumn{3}{c}{RtGender}                                                     \\ \hline
Method    & \multicolumn{1}{c}{\begin{tabular}[c]{@{}c@{}}Baseline\\ (None)\end{tabular}} & \multicolumn{1}{c}{CDA} & \multicolumn{1}{c|}{Dropout} & \multicolumn{1}{c}{None} & \multicolumn{1}{c}{CDA} & \multicolumn{1}{c|}{Dropout} & \multicolumn{1}{c}{None} & \multicolumn{1}{c}{CDA} & \multicolumn{1}{c}{Dropout} \\ \hline
BERT     & 57.25  & 55.34  & 55.73  & 51.91   & 42.37  & 48.09  & 56.11  & 47.71  & 41.98 \\
GPT2     & 56.87& 54.96 & 57.63  & 47.71  & 50.00  & 52.67  & 46.18  & 51.53& 47.33 \\ \hline
\end{tabular}
}
\caption{Stereotype scores tested on Crow-S. The lower the value, the more debiased the model is. The table represents the scores of models not fine-tuned, and of models fine-tuned on the  downstream task of toxicity detection, on Jigsaw and RtGender corpus respectively}
\label{tab:1}
\end{table*}

\begin{table*}[]
{\centering
\begin{tabular}{l|lllllllll}
\hline
Model& \multicolumn{9}{c}{BERT}\\ \hline
Finetuned& \multicolumn{3}{c|}{None}& \multicolumn{3}{c|}{Jigsaw}& \multicolumn{3}{c}{RtGender} \\ \hline
Debiasing method & None   & CDA    & \multicolumn{1}{l|}{Dropout} & None   & CDA    & \multicolumn{1}{l|}{Dropout} & None    & CDA     & Dropout  \\ \hline
SEAT 6& 0.931* & 0.785* & \multicolumn{1}{l|}{0.889*}  & 0.558* & 0.597* & \multicolumn{1}{l|}{0.515*}  & -0.268  & 1.963*  & 0.912*   \\
SEAT 6b& 0.089  & 0.083  & \multicolumn{1}{l|}{0.277}   & 0.169  & -0.104 & \multicolumn{1}{l|}{0.400*}  & 0.227   & 1.895*  & 0.391*   \\
SEAT 7& -0.124 & -0.512 & \multicolumn{1}{l|}{0.171}   & 1.035* & -0.626 & \multicolumn{1}{l|}{1.223*}  & 0.060   & 0.396*  & 0.351    \\
SEAT 7b& 0.936* & 1.238* & \multicolumn{1}{l|}{0.849*}  & 0.711* & 0.663* & \multicolumn{1}{l|}{1.135*}  & -0.085  & 0.506*  & 0.310    \\
SEAT 8& 0.782* & 0.025  & \multicolumn{1}{l|}{0.594*}  & 0.539* & -0.729 & \multicolumn{1}{l|}{0.551*}  & -0.091  & 0.786*  & 0.930*   \\
SEAT 8b& 0.858* & 0.673* & \multicolumn{1}{l|}{0.945*}  & 0.286  & 0.586* & \multicolumn{1}{l|}{0.600*}  & -0.205  & 0.817*  & 0.929*   \\ \hline
Model& \multicolumn{9}{c}{GPT2}\\ \hline
SEAT 6& 0.137  & 0.287  & \multicolumn{1}{l|}{0.288}   & 0.451* & 0.029  & \multicolumn{1}{l|}{0.667*}  & 1.359*  & 1.516*  & 1.554*   \\
SEAT 6b& 0.003  & 0.012  & \multicolumn{1}{l|}{0.032}   & 0.554* & 0.247  & \multicolumn{1}{l|}{0.418*}  & 0.893*  & 1.242*  & 0.976*   \\
SEAT 7& -0.023 & 0.862* & \multicolumn{1}{l|}{0.850*}  & 0.129  & 0.700* & \multicolumn{1}{l|}{0.751*}  & 1.044*  & -0.337  & 0.693*   \\
SEAT 7b& 0.001  & 0.933* & \multicolumn{1}{l|}{0.819*}  & 0.645* & 1.172* & \multicolumn{1}{l|}{1.041*}  & 1.060*  & -0.205  & 1.017*   \\
SEAT 8& -0.223 & 0.501* & \multicolumn{1}{l|}{0.486*}  & -0.057 & 0.545* & \multicolumn{1}{l|}{0.321}   & 0.867*  & -0.213  & 0.700*   \\
SEAT 8b& -0.286 & 0.278  & \multicolumn{1}{l|}{0.092}   & 0.059  & 0.222  & \multicolumn{1}{l|}{0.197}   & 0.783*  & -0.288  & 0.984*   \\ \hline
\end{tabular}}
\caption{The effect size of SEAT. The small effect size is an indication of the less biased model. * denotes the significance of p-value<0.01}
\label{tab:SEAT}
\end{table*}

\section{Results}
\subsection{Testing the efficacy of debiasing techniques}

\textbf{CrowS}  \hspace{1.5pt} Table \ref{tab:1} shows the debias stereotype scores across for debiasing methods on the CrowS dataset. We tested CrowS on two different models, BERT \textit{(bert-base-uncased)} and GPT2 \textit{(gpt2-small)}. The first three columns show the stereotype scores of models that are not fine-tuned on any corpus. We consider these models as baseline models. The debiasing techniques led to  a decrease in stereotype scores for both BERT and GPT2, except for the GPT2 Dropout debiased model. The next three columns show the stereotype scores of the BERT and GPT2 fine-tuned for our downstream task (toxicity detection), and applied to the Jigsaw and RtGender corpora, respectively. Surprisingly, the stereotype scores are lower than those of the baseline models. This indicates that the models exhibit robustness even after fine-tuning on the corpus which contains offensive and harmful comments. In fact, the results confirm the findings in \cite{webster2020measuring}, where CDA and Dropout debiasing methods showed \textit{resilience} to fine-tuning. However, this result needs extra investigation, as \cite{babaeianjelodar2020quantifying} suggesting that the BERT model fine-tuned on Jigsaw toxicity and RtGender, especially the latter, show an increase in direct gender bias measures compared to the baseline models.\\

\noindent\textbf{SEAT} \hspace{1.5pt} In order to check the generalizability of the debiasing effects, we calculated a different bias measure, SEAT \cite{may2019measuring}. Table \ref{tab:SEAT} shows the effect size of SEAT. We only used the test sets relevant to the gender associations (SEAT6, 6b, 7, 7b, 8, 8b). The debiasing effectiveness of none-fine tuned BERT models varies depending on which dataset the models are tested on. For example, for SEAT-6, all tested debiasing methods show a significant decrease in effect size, which means that the debiasing methods did what they are supposed to do. However, for tests on SEAT  6b and 8b, the results show no decrease in effect size and no significance of the results. Interestingly, the degree of effectiveness varies based on which corpus a model is fine-tuned. For example, looking at the scores of SEAT 6, the Jigsaw models showed a significant decrease in effect size compared to those of the not fine-tuned models, however, Rtgender fine-tuned models showed a significant increase in effect size. These outcomes further support the findings by \cite{babaeianjelodar2020quantifying}, i.e., that because the Jigsaw dataset involves comments related to \textit{race} and \textit{sexuality} rather than gender, the gender bias learned from the corpus is less severe than for RtGender.
\begin{table*}[!ht]
\centering
\begin{tabular}{l|lll|lll}
\hline
Model& \multicolumn{3}{c|}{BERT}& \multicolumn{3}{c}{GPT2}\\ \hline
Methods  & None  & CDA& Dropout& None  & CDA& Dropout \\ \hline
Jigsaw   & 0.944 & 0.919& \textbf{0.949} & 0.950 & 0.929& 0.947\\
RtGender & 0.570 & \textbf{0.747} & 0.558& 0.698 & \textbf{0.716} & 0.703\\ \hline
\end{tabular}
\caption{Accuracy score of \textit{toxicity detection task} Jigsaw and RtGender respectively}
\label{tab:Acc}
\end{table*}

\begin{table*}[]
\centering
\begin{tabular}{l|llllll}
\hline
& Baseline & CDA    & Dropout & Jigsaw & \begin{tabular}[c]{@{}l@{}}Jigsaw\\ CDA\end{tabular} & \begin{tabular}[c]{@{}l@{}}Jigsaw\\ Dropout\end{tabular} \\ \hline
Total effect& 2.865    & 2.046  & 1.858   & 0.122  & 0.116& 0.092\\
Male total effect   & 3.964    & 2.792  & 2.514   & 0.122  & 0.116& 0.092\\
Female total effect & 30.227   & 25.953 & 23.550  & 0.752  & 0.979& 0.502\\ \hline
\end{tabular}
\caption{Total effect statistics.}
\label{tab:TE}
\end{table*}
The results are less clear for GPT2. Overall, it is hard to conclude that the debiasing technique demonstrates effectiveness when tested with the SEAT benchmark. Looking only at the results that show significance (p-value<0.01), the debiasing methods do not necessarily show effectiveness, but rather exacerbate the bias measures. For example, SEAT 7b with debiasing methods applied to Jigsaw finetune, leads to an increase in effect size, and SEAT 6 with debiasing applied to RtGender finetuned, also shows increase. One of the reasons for this observation could be that SEAT measures association of gendered \textit{names} with professions, while debiasing methods focus on gendered pronouns, not on the gender of a name. Overall, our results suggest that testing the bias of language models on a single \textit{bias} measure may not be reliable enough as measures may differ across models and corpora on which language models are fine-tuned. This may be in part due to the fact that \textit{gender bias} is an inherently complex concept that furthermore depends on contexts of text production and use, and how "gender" it is defined and measured. Thus, evaluation on two or more benchmark datasets is desirable.\\

\textbf{Accuracy} \hspace{1.5pt}  Table \ref{tab:Acc} shows the accuracy scores of the models on downstream task, \textit{toxicity detection}. Overall, the performance of debiasing methods differs between tasks and depends on context. This supports the findings in \cite{meade2021empirical}. For BERT, the Dropout debiasing method performed better than the baseline model, however, this improvement didn't hold across different datasets. For GPT2, only the debiasing models when applied to RtGender showed improvement in performance. 

\subsection{Causal Mediation Analysis}

\begin{figure}[]
\centerline{\includegraphics[width=3in, height=1.5in]{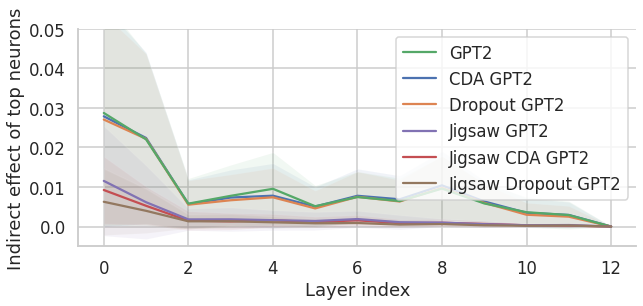}}
\caption{The indirect effect of the top neurons by layer index.}
\label{fig:layer_index}
\end{figure}

\textbf{Total Effect} \hspace{1.5pt} Table \ref{tab:TE} shows the total effect across models. Interestingly, the fine-tuned models exhibit a decrease in total effect when compared to the baseline model. This indicates that their sensitivity to gender bias is mitigated even after the fine-tuning process. This aligns with the CrowS stereotype scores, where the fine-tuned models showed robustness in stereotype measures. Besides the total effect, the male and female total effect was measured by splitting the profession dataset \cite{bolukbasi2016man} based on stereotypical male and female professions, respectively. The results show that the effect size is higher for female cases, which means that the language model exhibits more sensitivity for female professions. According to \cite{vig2020investigating}, this may be in part due to the fact the stereotypes related to professions of females are stronger than those related to males.\\

\begin{figure*}[t]
\centerline{\includegraphics[width=7.5in, height=5in]{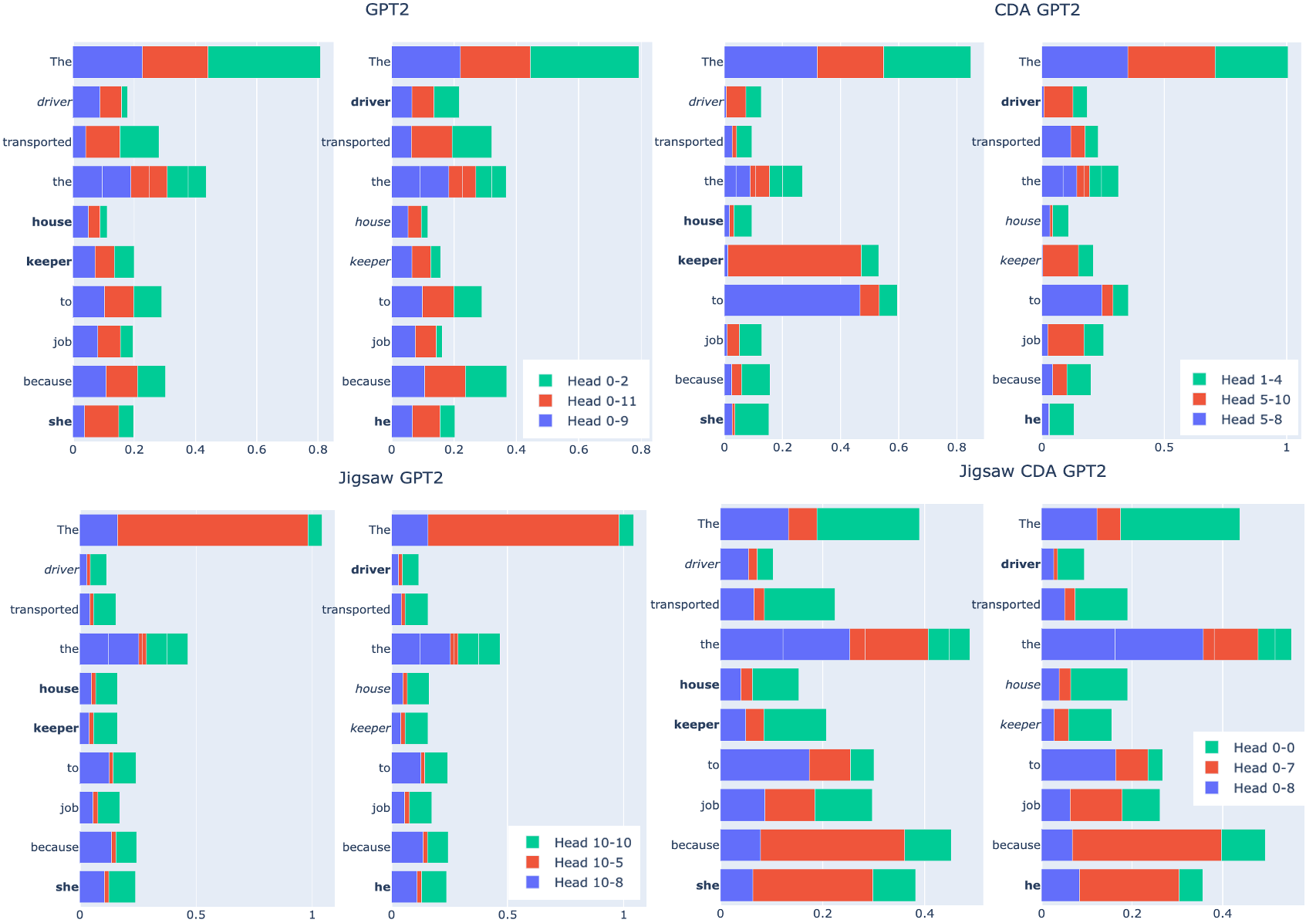}}
\caption{Weights distribution of the top attention heads of the models on two different prompts. The labels indicate the layer-attention head index. For example, Head 0-2 refers to attention head index 2, in layer 0.}
\label{fig:attention}
\end{figure*}

\noindent\textbf{Neurons interventions} \hspace{1.5pt}  Figure \ref{fig:layer_index} shows the indirect effect distribution of the top 2.5\% of the neurons. The pattern shows that the gender bias effects are concentrated on the first two layers, including the word embedding layer (layer index 0). Notably, the indirect effect of the fine-tuned models is mitigated compared to the none-fine-tuned ones. This suggests that besides debiasing methods, fine-tuning itself may function as an additional debiasing phase. Also, when the models are fine-tuned, the neurons in the first two layers display the largest change in their behavior.
\\

\noindent\textbf{Attention head interventions} \hspace{1.5pt} Figure \ref{fig:attention} shows a qualitative analysis of the attention head interventions. The figure presents the distribution of the attention weights of the top 3 attention heads, given the two different sentences `The driver transported the \textbf{housekeeper} to job because \textbf{she}’ and ‘The \textbf{driver} transported the housekeeper to job because \textbf{he}’. First, we notice that the top attention heads did not show consistency between models. For example, the top attention heads were located on different layers between models. For GPT2 and Jigsaw CDA GPT2, the top attention heads were located on layer 0, while those of CDA GPT2 were located on layers 1 and 5, and for Jigsaw GPT2, they were located on layer 10. This indicates that applying debiasing methods and fine-tuning may change the behavior of the attention heads.\\

Second, the debiased models (e.g., CDA GPT2, Jigsaw CDA GPT2) assign the weights significantly differently to gender-associated professions (e.g., driver, housekeeper). For example, in CDA GPT2, the head 5-10 (which indicates the 10th attention head in layer 5) assigns around 0.5 to the word ‘keeper’ in the first plot, while it attends around 0.2 to that of the second plot. The head 5-10 in CDA GPT2 also attends around 0.1 to the word ‘driver’ in the first plot, while assigning more than 0.1 to the ‘driver’ in the second plot. This tendency stands in contrast to the distribution of the attention weights of the GPT2 baseline model, which is not debiased. Such changes in attention weights in gender-associated terms may indicate that debiasing and fine-tuning methods may modify the behavior of the attention heads, suggesting the model what to \textit{be aware of}.\\

\section{Conclusion}
\label{sec:conclusion}
In this work, we have investigated how debiasing methods impact language models, along with the downstream tasks. We found that (1) debiasing methods are robust after fine-tuning on downstream tasks. In fact, after the fine-tuning, the debiasing effects strengthened. However, this effect is not supported across another bias measure. This indicates the need for both debiasing techniques and bias benchmarks to ensure generalizability. The causal mediation analysis suggests that (2) The neurons that showed a large change in behavior were located in the first two layers of language models (including the word embedding layers). This suggests that careful inspection of certain components of the language models is recommended when applying debiasing methods. (3) Applying debiasing and fine-tuning methods to language models changes the weight that attention heads assign to gender-associated terms. This indicates that attention heads may play a crucial role in representing gender bias in language models.  \\Several limitations apply to this work. We only tested these effects on one downstream task, namely, toxicity detection. In order to check the generalizability of these findings, experiments with other downstream tasks are necessary.

\bibliography{naacl2021}
\bibliographystyle{acl_natbib}

\clearpage
\appendix

\section{Appendix}
\label{sec:appendix}
\begin{figure}[H]
\centering
\begin{flushright}
\includegraphics[width=18cm,height=23cm,keepaspectratio]{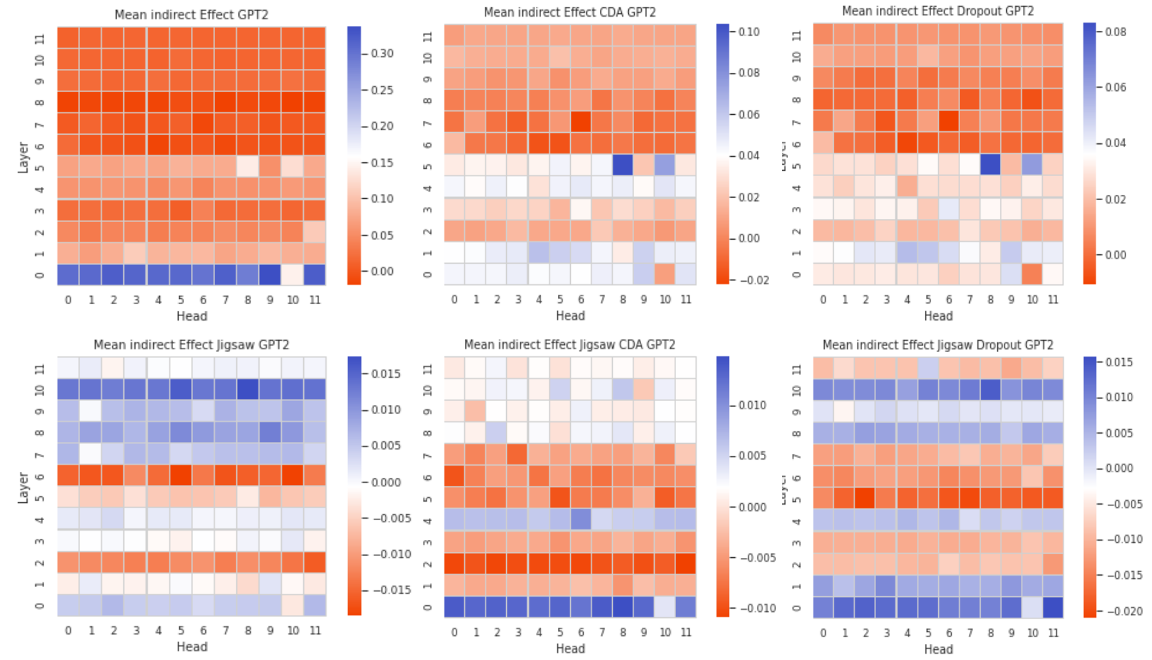}
\caption{Main indirect Effect of attention intervention.}
\label{fig:mean_indirect}
\end{flushright}
\end{figure}

\begin{figure*}[!htbp]
\centerline{\includegraphics{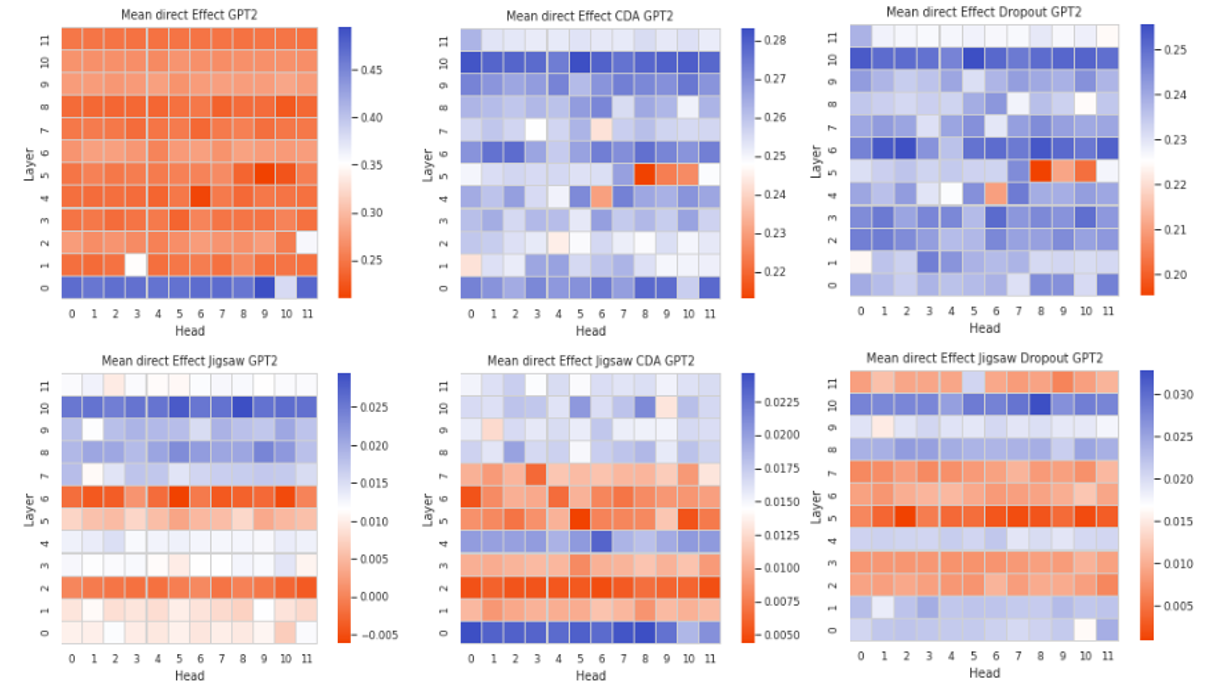}}
\caption{Main direct Effect of Attention Intervention}
\label{fig:mean_direct}
\end{figure*}

\begin{figure*}[!htbp]
\centerline{\includegraphics[width=4.5in, height=4in]{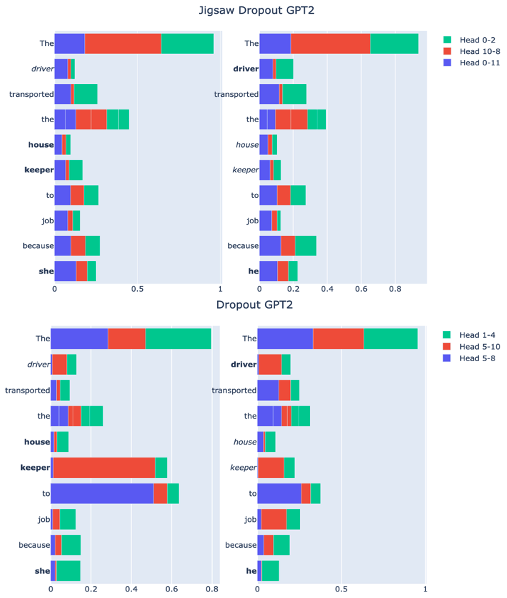}}
\caption{Attention weights of Dropout debiased models}
\label{fig:attn_dropout}
\end{figure*}
\end{document}